\pdfoutput=1
\documentclass{article}
\usepackage[dvipsnames]{xcolor}
\usepackage{tikz}
\usepackage{hyperref}
\usepackage{cleveref}[2012/02/15]
\usetikzlibrary{shapes,arrows,calc}
\tikzstyle{decision} = [diamond, draw, fill=RoyalPurple!15, 
    text width=6em, text centered, node distance=3cm, inner sep=2pt]
\tikzstyle{block} = [rectangle, draw, fill=YellowGreen!15, 
    text width=10em, text centered, rounded corners, minimum height=3em]
\tikzstyle{block2} = [rectangle, draw, fill=RoyalBlue!15, 
    text width=5em, text centered, rounded corners, minimum height=2em]
\tikzstyle{block3} = [rectangle, draw, fill=Gray!10, 
text width=10em, text centered, rounded corners, minimum height=3em]
\tikzstyle{block4} = [rectangle, draw, fill=RoyalBlue!15, 
    text width=8em, text centered, rounded corners, minimum height=2em]
\tikzstyle{block5} = [rectangle, draw, fill=RoyalBlue!15, 
    text width=7em, text centered, rounded corners, minimum height=3.5em]
\tikzstyle{cloud1} = [cloud, draw,cloud puffs=10,cloud puff arc=120, aspect=2, inner ysep=0.5em, fill=Yellow!20,]
\tikzstyle{line} = [draw, -latex']
\crefformat{footnote}{#2\footnotemark[#1]#3}
\makeatletter
\pgfdeclareshape{document}{
\inheritsavedanchors[from=rectangle] 
\inheritanchorborder[from=rectangle]
\inheritanchor[from=rectangle]{center}
\inheritanchor[from=rectangle]{north}
\inheritanchor[from=rectangle]{south}
\inheritanchor[from=rectangle]{west}
\inheritanchor[from=rectangle]{east}
\backgroundpath{
\southwest \pgf@xa=\pgf@x \pgf@ya=\pgf@y
\northeast \pgf@xb=\pgf@x \pgf@yb=\pgf@y
\pgf@xc=\pgf@xb \advance\pgf@xc by-10pt 
\pgf@yc=\pgf@yb \advance\pgf@yc by-10pt
\pgfpathmoveto{\pgfpoint{\pgf@xa}{\pgf@ya}}
\pgfpathlineto{\pgfpoint{\pgf@xa}{\pgf@yb}}
\pgfpathlineto{\pgfpoint{\pgf@xc}{\pgf@yb}}
\pgfpathlineto{\pgfpoint{\pgf@xb}{\pgf@yc}}
\pgfpathlineto{\pgfpoint{\pgf@xb}{\pgf@ya}}
\pgfpathclose
\pgfpathmoveto{\pgfpoint{\pgf@xc}{\pgf@yb}}
\pgfpathlineto{\pgfpoint{\pgf@xc}{\pgf@yc}}
\pgfpathlineto{\pgfpoint{\pgf@xb}{\pgf@yc}}
\pgfpathlineto{\pgfpoint{\pgf@xc}{\pgf@yc}}
}
}
\makeatother

\usepackage{bera}
\usepackage{listings}
\colorlet{punct}{red!60!black}
\definecolor{background}{HTML}{f9f9f9}
\definecolor{delim}{RGB}{20,105,176}
\colorlet{numb}{magenta!80!black}

\lstdefinelanguage{json}{
    basicstyle=\normalfont\ttfamily,
    numberstyle=\scriptsize,
    stepnumber=1,
    numbersep=8pt,
    showstringspaces=false,
    breaklines=true,
    frame=lines,
    backgroundcolor=\color{background},
    literate=
     *{0}{{{\color{numb}0}}}{1}
      {1}{{{\color{numb}1}}}{1}
      {2}{{{\color{numb}2}}}{1}
      {3}{{{\color{numb}3}}}{1}
      {4}{{{\color{numb}4}}}{1}
      {5}{{{\color{numb}5}}}{1}
      {6}{{{\color{numb}6}}}{1}
      {7}{{{\color{numb}7}}}{1}
      {8}{{{\color{numb}8}}}{1}
      {9}{{{\color{numb}9}}}{1}
      {:}{{{\color{punct}{:}}}}{1}
      {,}{{{\color{punct}{,}}}}{1}
      {\{}{{{\color{delim}{\{}}}}{1}
      {\}}{{{\color{delim}{\}}}}}{1}
      {[}{{{\color{delim}{[}}}}{1}
      {]}{{{\color{delim}{]}}}}{1},
}

\usepackage{amsfonts}
\usepackage{microtype}
\usepackage{graphicx}
\usepackage{subfigure}
\usepackage{booktabs} 
\graphicspath{ {./images/} }

\usepackage{hyperref}
\usepackage{multirow}

\usepackage[accepted]{sysml2019}

\sysmltitlerunning{PyText: A seamless path from NLP research to production using PyTorch}

\newcommand{\mrinal}[1]{}
\renewcommand{\mrinal}[1]{{\color{red} [Mrinal: {#1}]}}
\newcommand{\sonal}[1]{}
\renewcommand{\sonal}[1]{{\color{red} [Sonal: {#1}]}}
\newcommand{\abhinav}[1]{}
\renewcommand{\abhinav}[1]{{\color{red} [Abhinav: {#1}]}}

\newcommand{\tick}[0]{\textcolor{green}{\checkmark}}
\newcommand{\cross}[0]{\textcolor{red}{$\times$}}
\newcommand*{\affmark}[1][*]{\textsuperscript{#1}}
\newcommand*{\affaddr}[1]{#1}

\begin{document}

\twocolumn[
\sysmltitle{PyText: A seamless path from NLP research to production}




\begin{sysmlauthorlist}
\sysmlauthor{Ahmed Aly}{fbconv}
\sysmlauthor{Kushal Lakhotia}{fbconv}
\sysmlauthor{Shicong Zhao}{fbconv}
\sysmlauthor{Mrinal Mohit}{fbconv}
\sysmlauthor{Barlas O\u{g}uz}{fbaml}
\sysmlauthor{Abhinav Arora}{fbconv}
\sysmlauthor{Sonal Gupta}{fbconv}
\sysmlauthor{Christopher Dewan}{fbaml}
\sysmlauthor{Stef Nelson-Lindall}{fbaml}
\sysmlauthor{Rushin Shah}{fbconv}
\end{sysmlauthorlist}

\sysmlaffiliation{fbconv}{Facebook Conversational AI}
\sysmlaffiliation{fbaml}{Facebook Applied Machine learning}

\sysmlkeywords{Machine Learning, NLP, AI, NLU}

\vskip 0.3in
\vspace{-2em}
\begin{center}
\affaddr{\affmark[1]Facebook Conversational AI}\\
\affaddr{\affmark[2]Facebook AI}\\
\end{center}
\vskip 0.3in
\begin{abstract}
We introduce PyText\footnote{https://github.com/facebookresearch/pytext/} -- a deep learning based NLP modeling framework built on PyTorch. PyText addresses the often-conflicting requirements of enabling rapid experimentation and of serving models at scale. It achieves this by providing simple and extensible interfaces for model components, and by using PyTorch's capabilities of exporting models for inference via the optimized Caffe2 execution engine. We report our own experience of migrating experimentation and production workflows to PyText, which enabled us to iterate faster on novel modeling ideas and then seamlessly ship them at industrial scale.

\end{abstract}
]

\section{Introduction}
\label{introduction}

When building a machine learning system, especially one based on neural networks, there is usually a trade-off between ease of experimentation and deployment readiness, often with conflicting requirements. For instance, to rapidly try out flexible and non-conventional modeling ideas, researchers tend to use modern imperative deep-learning frameworks like PyTorch\footnote{https://pytorch.org/} or TensorFlow Eager\footnote{https://www.tensorflow.org/guide/eager}. These frameworks provide an easy, eager-execution interface that facilitates writing advanced and dynamic models quickly, but also suffer from overhead in latency at inference and impose deployment challenges. In contrast, production-oriented systems are typically written in declarative frameworks that express the model as a static graph, such as Caffe2\footnote{https://caffe2.ai/} and Tensorflow\footnote{https://www.tensorflow.org/}. While being highly optimized for production scenarios, they are often harder to use, and make the experimentation life-cycle much longer. This conflict is even more prevalent in natural language processing (NLP) systems, since most NLP models are inherently very dynamic, and not easily expressible in a static graph. This adds to the challenge of serving these models at an industrial scale.

PyText, built on PyTorch 1.0~\footnote{https://pytorch.org/blog/the-road-to-1\_0/}, is designed to achieve the following:
\begin{enumerate}
\item Make experimentation with new modeling ideas as easy and as fast as possible.
\item Make it easy to use pre-built models on new data with minimal extra work.
\item Define a clear workflow for both researchers and engineers to build, evaluate, and ship their models to production with minimal overhead.
\item Ensure high performance (low latency and high throughput) on deployed models at inference.
\end{enumerate}

\begin{table}[h]
\begin{tabular}{|l|c|c|c|}
\hline
\textbf{\begin{tabular}{@{}c@{}}NLP\\Framework\end{tabular}} & \textbf{\begin{tabular}{@{}c@{}}Deep Learning\\Support\end{tabular}} & \textbf{\begin{tabular}{@{}c@{}}Easy\\Prototyping\end{tabular}} & \textbf{\begin{tabular}{@{}c@{}}Industrial\\Performance\end{tabular}} \\ \hline
CoreNLP              & \cross   & \tick     & \tick     \\ 
AllenNLP             & \tick    & \tick     & \cross    \\ 
FLAIR                & \tick    & \tick     & \cross    \\ 
Spacy 2.0            & \tick    & \cross    & \tick     \\ 
\textbf{PyText}      & \tick    & \tick     & \tick     \\ \hline
\end{tabular}
\caption{Comparison of NLP Modeling Frameworks}
\label{table:benchmark}
\end{table}

Existing popular frameworks for building state-of-the-art NLP models include Stanford CoreNLP~\cite{corenlp}, AllenNLP~\cite{Gardner2017AllenNLP}, FLAIR~\cite{akbik2018coling} and Spacy 2.0~\footnote{http://spacy.io}. CoreNLP has been a popular library for both research and production, but does not support neural network models very well. AllenNLP and FLAIR are easy-to-use for prototypes but it is hard to productionize the models since they are in Python, which doesn't support large scale real time requests due to lack of good multi-threading support. Spacy 2.0 has some state-of-the-art NLP models built for production use-cases but is not easily extensible for quick prototyping and building new models.

\section{Framework Design}
PyText is a modeling framework that helps researchers and engineers build end-to-end pipelines for training or inference. Apart from workflows for experimentation with model architectures, it provides ways to customize handling of raw data, reporting of metrics, training methodology and exporting of trained models. PyText users are free to implement one or more of these components and can expect the entire pipeline to work out of the box. A number of default pipelines are implemented for popular tasks which can be used as-is. We now dive deeper into building blocks of the framework and its design.

\subsection{Component}
Everything in PyText is a component. A component is clearly defined by the parameters required to configure it. All components are maintained in a global registry which makes PyText aware of them. They currently include --

\textbf{Task:} combines various components required for a training or inference task into a pipeline. Figure \ref{fig:doc_classification_jobspec} shows a sample config for a document classification task. It can be configured as a JSON file that defines the parameters of all the children components.

\textbf{Data Handler:} processes raw input data and prepare batches of tensors to feed to the model.

\textbf{Model:} defines the neural network architecture.

\textbf{Optimizer:} encapsulates model parameter optimization using loss from forward pass of the model.

\textbf{Metric Reporter:} implements the relevant metric computation and reporting for the models.

\textbf{Trainer:} uses the data handler, model, loss and optimizer to train a model and perform model selection by validating against a holdout set.

\textbf{Predictor:} uses the data handler and model for inference given a test dataset.

\textbf{Exporter:} exports a trained PyTorch model to a Caffe2 graph using ONNX\footnote{https://onnx.ai/}.

\begin{figure}[!h]
\centering
\scriptsize
\begin{lstlisting}[language=Json]
{
  "config": {
    "task": {
      "DocClassificationTask": {
        "data_handler": {
          "columns_to_read": ["doc_label", "text"],
          "shuffle": true
        },
        "model": {
          "representation": {
            "BiLSTMPooling": {
              "pooling": {
                "SelfAttention": {
                  "attn_dimension": 128,
                  "dropout": 0.4
                }
              },
              "bidirectional": true,
              "dropout": 0.4,
              "lstm": { "lstm_dim": 200, "num_layers": 2 }
            }
          },
          "output_config": {
            "loss": { "CrossEntropyLoss": {} }
          },
          "decoder": { "hidden_dims": [128] }
        },
        "features": {
          "word_feat": {
            "embed_dim": 200,
            "pretrained_embeddings_path": "/tmp/embeds",
            "vocab_size": 250000,
            "vocab_from_train_data": true
          }
        },
        "trainer": {
          "random_seed": 0,
          "epochs": 15,
          "early_stop_after": 0,
          "log_interval": 1,
          "eval_interval": 1,
          "max_clip_norm": 5
        },
        "optimizer": {
          "type": "adam",
          "lr": 0.001,
          "weight_decay": 0.00001
        },
        "metric_reporter": {
          "output_path": "/tmp/test_out.txt"
        },
        "exporter": {}
      }
    }
  }
}
\end{lstlisting}
\caption{Document Classification Task Config}
\label{fig:doc_classification_jobspec}
\end{figure}

\subsection{Design Overview}
The task bootstraps a PyText job and creates all the required components. There are two modes in which a job can be run:


\begin{figure}[!h]
\begin{center}
\small
\begin{tikzpicture}[node distance = 1cm, auto]
    \node [block2] (init) {Data Handler};
    \node [block2, below of=init, node distance=1.75cm] (trainer) {Trainer};
    \node [block2, left of=trainer, node distance=2.75cm] (task) {Task};
    \node [block2, right of=trainer, node distance=2.75cm] (exporter) {Exporter};
    \node [cloud1, below of=exporter, node distance=2cm, text width=1cm] (caffe2) {Caffe2 Model};
    \node [block3, below of=trainer, node distance=3.5cm, text width=2cm, minimum height=4.25cm] (components) {};
    \node [block2, below of=trainer, node distance=2.0cm] (model) {Model};
    \node [block2, below of=model] (loss) {Loss};
    \node [block2, below of=loss] (optimizer) {Optimizer};
    \node [block2, below of=optimizer] (metrics) {Metric Reporter};
    \path [line] (init) -- node[text width=1.25cm, text centered, left=0.15cm] {Batch Iterators} (trainer);
    \path [line] (components) -- node[text width=1.25cm, left=0.35cm] {Components} (trainer);
    \path [line] (task) |- (init);
    \path [line] (task) |- (components);
    \path [line] (task) -- (trainer);
    \path [line] (trainer) -- node[text width=1.25cm, text centered, above=.25cm] {PyTorch Model} (exporter);
    \path [line] (exporter) -- (caffe2);
\end{tikzpicture}
\caption{PyText Framework Design}
\label{fig:PyTextdesign}
\end{center}
\end{figure}
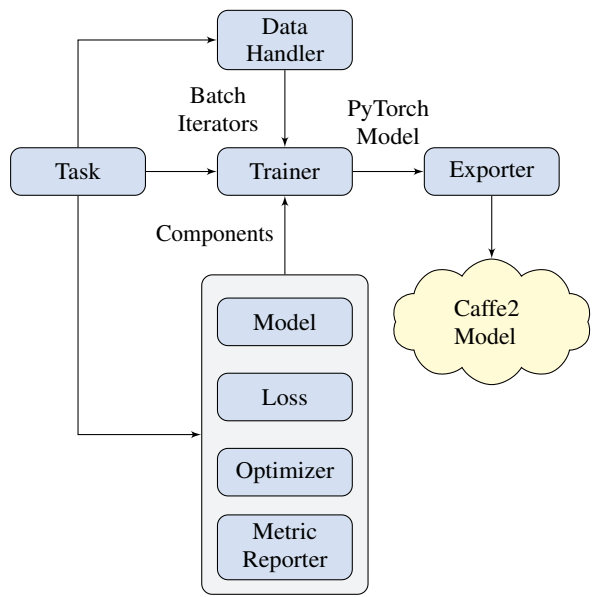

\begin{itemize}
    \item \textit{Train}: Trains a model either from scratch or from a saved check-point. Task uses the Data Handler to create batch iterators over training, evaluation and test data-sets and passes these iterators along with model, optimizer and metrics reporter to the trainer. Subsequently, the trained model is serialized in PyTorch format as well as converted to a static Caffe2 graph.
    \item \textit{Predict}: Loads a pre-trained model and computes its prediction for a given test set. The task Manager, again, uses the Data Handler to create a batch iterator over the test data-set and passes it with the model to the predictor for inference.
\end{itemize}
Figure \ref{fig:PyTextdesign} illustrates the overall design of the framework.
\section{Modeling Support}
We now discuss the native support for building and extending models in PyText.
\subsection{Terminology}
\textbf{Module:} is a reusable component that is implemented without any knowledge of which model it will be used in. It defines a clear input and output interface such that it can be plugged into another module or model.

\textbf{Model:} has a one-to-one mapping with a task. Each model can be made up of a combination of modules for running a training or prediction job.

\subsection{Model Abstraction} \label{modelabstraction}
PyText provides a simple, easily extensible model abstraction. We break up a single-task model into Token Embedding, Representation, Decoder and Output layers, each of which is configurable. Further, each module can be saved and loaded individually to be reused in other models.

\textbf{Token Embedding:} converts a batch of numericalized tokens into a batch of vector embeddings for each token. It can be configured to use embeddings of a number of styles: pre-trained word-based, trainable word-based, character-based with CNN and highway networks\cite{DBLP:conf/aaai/KimJSR16}, pre-trained deep contextual character-based (e.g., ELMo\cite{N18-1202}), token-level gazetteer features or morphology-based (e.g. capitalization).

\textbf{Representation:} processes a batch of embedded tokens to a representation of the input. The implementation of what it emits as output depends on the task, e.g., the representation of the document for a text classification task will differ from that for a word tagging task. Logically this part of the model should implement the sub-network such that its output can be interpreted as features over the input. Examples of the different representations that are present in PyText are; Bidirectional LSTM and CNN representations.

\textbf{Decoder:} is responsible for generating logits from the input representation. Logically this part of the model should implement the sub-network that generates model output over the features learned by the representation.

\textbf{Output Layer:} concerns itself with generating prediction and the loss (when label or ground truth is provided).

These modules compose the base model implementation, they can be easily extended for more complicated architectures.

\subsection{Multi-task Model Training}
PyText supports multi-task training \cite{collobert:icml08} to optimize multiple tasks jointly as a first-class citizen. We use multi-task model by allowing parameter sharing between modules of the multiple single task models. We use the model abstraction for single task discussed in Section \ref{modelabstraction} to define the tasks and let the user declare which modules of those single tasks should be shared. This enables training a model with one or more input representations jointly against multiple tasks.

Multi-task models make the following assumptions:
\begin{itemize}
  \item If there are n tasks in the multi-task model setup then there must be n data sources containing data for one task each.
  \item The single task scenario must be implemented for it to be reused for the multi-task setup.
\end{itemize}


\begin{figure}[!h]
\begin{center}
\small
\begin{tikzpicture}[node distance = 1.75cm, auto]
    \node [block5] (init) {Embedding};
    \node [block5, below of=init] (rep) {Document\\Representation};
    \node [block5, below of=rep, xshift=-2.5cm] (docdecoder) {Document Classification Decoder};
    \node [block5, below of=rep, xshift=2.5cm] (worddecoder) {Word Tagging Decoder};
    \node [block5, below of=docdecoder] (docoutput) {Classification Output Layer};
    \node [block5, below of=worddecoder] (wordoutput) {Word Tagging Output Layer};
    \path [line] (init) -- (rep);
    \path [line] (rep.west) -| (docdecoder.north);
    \path [line] (rep.east) -| (worddecoder.north);
    \path [line] (docdecoder) -- (docoutput);
    \path [line] (worddecoder) -- (wordoutput);
\end{tikzpicture}
\caption{Joint document classification and word tagging model}
\label{fig:DocMultitask}
\end{center}
\end{figure}
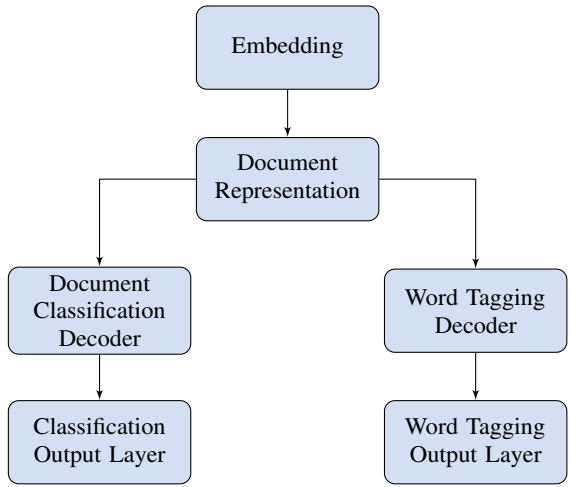

\subsubsection{Multi-task Model Examples}
PyText provides the flexibility of building any multi-task model architecture with the appropriate model configuration, if the two assumptions listed above are satisfied. The examples below give a flavor of two sample model architectures built with PyText for joint learning against more than one task.

\textbf{Figure \ref{fig:DocMultitask}} illustrates a model that learns a shared document representation for document classification and word tagging tasks. This model is useful for natural language understanding where given a sentence, we want to predict the intent behind it and tag the slots in the sentence. Jointly optimizing for two tasks helps the model learn a robust sentence representation for the two tasks. Further, we can use this pre-trained sentence representation for other tasks where training data is scarce.


\begin{figure}[!h]
\begin{center}
\small
\begin{tikzpicture}[node distance = 1.75cm, auto]
    \node [block5] (init) {Embedding};
    \node [block5, below of=init, xshift=-3cm] (docrep) {Document Representation};
    \node [block5, below of=init, xshift=3cm] (queryrep) {Query\\Representation};
    \node [block5, below of=docrep] (docdecoder1) {Document Classification Decoder};
    \node [block5, below of=queryrep] (docdecoder2) {Document Classification Decoder};
    \node [block5, right of=docdecoder1, node distance = 3cm] (querysim) {Query Similarity Decoder};
    \node [block5, below of=docdecoder1] (docoutput1) {Classification Output Layer};
    \node [block5, below of=docdecoder2] (docoutput2) {Classification Output Layer};
    \node [block5, below of=querysim] (docoutput3) {Classification Output Layer};
    \path [line] (init.west) -| (docrep.north);
    \path [line] (init.east) -| (queryrep.north);
    \path [line] (docrep.east) -| ($(querysim.north)+(-0.2,0)$);
    \path [line] (queryrep.west) -| ($(querysim.north)+(0.2,0)$);
    \path [line] (docrep) -- (docdecoder1);
    \path [line] (docdecoder1) -- (docoutput1);
    \path [line] (queryrep) -- (docdecoder2);
    \path [line] (docdecoder2) -- (docoutput2);
    \path [line] (querysim) -- (docoutput3);
\end{tikzpicture}
\caption{Joint query-document relevance and document classification model}
\label{fig:QueryDocMultitask}
\end{center}
\end{figure}
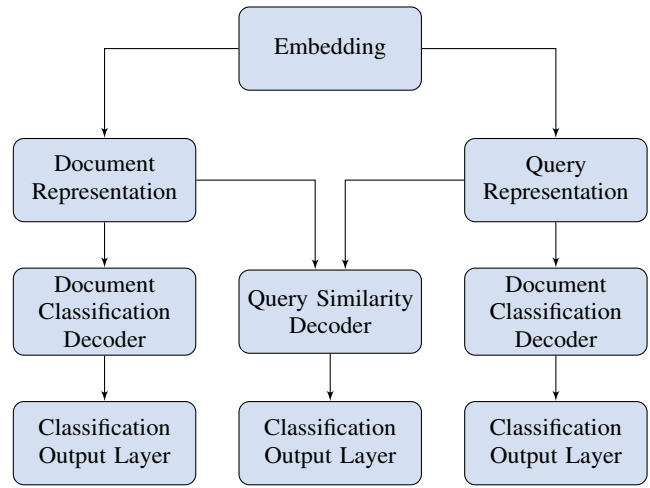

\textbf{Figure \ref{fig:QueryDocMultitask}} illustrates a model that learns document and query representations using query-document relevance and individual query and document classification tasks. This is often used in information retrieval where, given a query and a document, we want to predict their relevance; but we also add query and document classification tasks to increase robustness of learned representations.

\subsection{Model Zoo}
PyText models are focused on NLP tasks that can be configured with a variety of modules. We enumerate here the classes of models that are currently supported. 
\begin{itemize}
    \item Text Classification: classifies a sentence or a document into an appropriate category. PyText includes reference implementations of Bidirectional LSTM \cite{Schuster:1997:BRN:2198065.2205129} with Self-Attention \cite{lin+al-2017-embed-iclr} and Convolutional Neural Network \cite{D14-1181} models for text classification.
    \item Word Tagging: labels word sequences, i.e. classifies each word in a sequence to an appropriate category. Common examples of such tasks include Part-of-Speech (POS) tagging, Named Entity Recognition (NER) and Slot Filling in spoken language understanding. PyText contains reference implementations of Bidirectional LSTM with Slot-Attention and Bidirectional Sequential Convolutional Neural Network \cite{Vu2016} for word tagging. 
    \item Semantic Parsing: maps a natural language sentence into a formal representation of its meaning. PyText provides a reference implementation for Recurrent Neural Network Grammars \cite{DBLP:conf/naacl/DyerKBS16} \cite{RNNGEMNLP2018} for semantic parsing.
    \item Language Modeling: assigns a probability to a sequence of words (sentence) in a language. It also assigns a probability for the likelihood of a given word to follow a sequence of words. PyText provides a reference implementation for a stacked LSTM Language Model \cite{DBLP:conf/interspeech/MikolovKBCK10}. 
    \item Joint Models: We utilize the multi-task training support illustrated earlier to fuse and train models for two or more of the tasks mentioned here and optimize their parameters jointly.
\end{itemize}

\section{Production Workflow}
\subsection{From Idea to Production}
Researchers and engineers can follow the following steps to validate their ideas and quickly ship them to production --


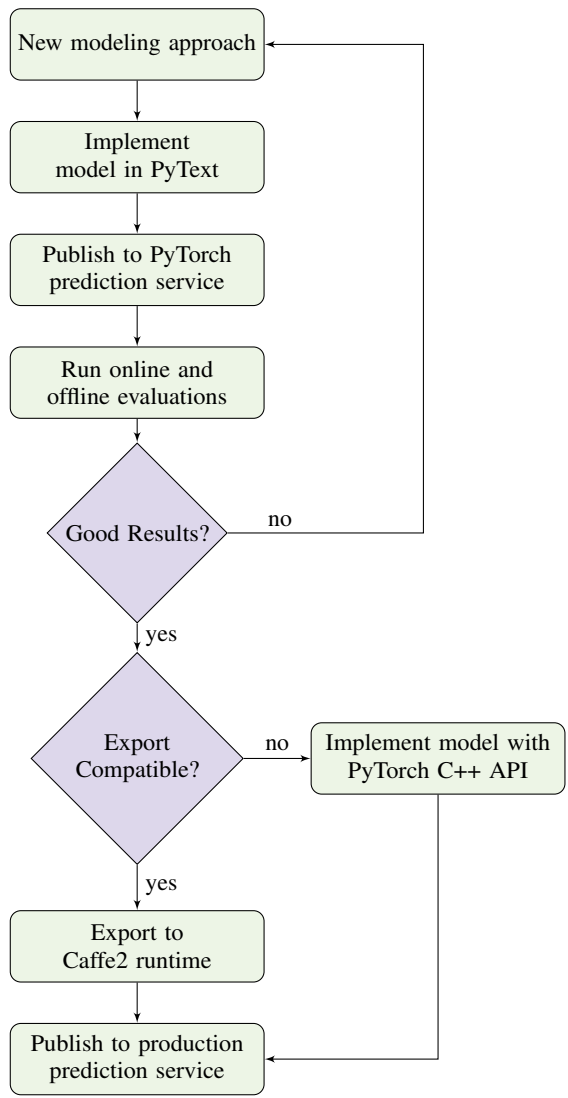
\begin{figure}[!h]
\begin{center}
\small
\begin{tikzpicture}[node distance = 1.5cm, auto]
    \node [block] (init) {New modeling approach};
    \node [block, below of=init] (implementpy) {Implement model in PyText};
    \node [block, below of=implementpy] (publishpy) {Publish to PyTorch prediction service};
    \node [block, below of=publishpy] (eval) {Run online and offline evaluations};
    \node [decision, below of=eval, node distance=2cm] (isgood) {Good Results?};
    \node [decision, below of=isgood] (canexport) {Export Compatible?};
    \node [block, right of=canexport, node distance=4cm] (cpp) {Implement model with PyTorch C++ API};
    \node [block, below of=canexport, node distance=2.5cm] (export) {Export to Caffe2 runtime};
    \node [block, below of=export] (publish) {Publish to production prediction service};
    \path [line] (init) -- (implementpy);
    \path [line] (implementpy) -- (publishpy);
    \path [line] (publishpy) -- (eval);
    \path [line] (eval) -- (isgood);
    \path [line] (isgood.east) --++(2.6,0) -- ++(0,4.5) -- ++(0,2) -- node [xshift=-0.85cm, yshift=-6.15cm] {no} (init.east);
    \path [line] (isgood) -- node {yes} (canexport);
    \path [line] (canexport) -- node {no} (cpp);
    \path [line] (canexport) -- node {yes} (export);
    \path [line] (cpp) |- (publish);
    \path [line] (export) -- (publish);
\end{tikzpicture}
\caption{From Idea to Production flowchart}
\label{fig:prod_chart}
\end{center}
\end{figure}

\begin{enumerate}
\item Implement the model in PyText, and make sure offline metrics on the test set look good.
\item Publish the model to the bundled PyTorch-based inference service, and do a real-time small scale evaluation on a live traffic sample.
\item Export it automatically to a Caffe2 net. In some cases, e.g. when using complex control flow logic and custom data-structures, this might not yet be supported via PyTorch 1.0.
\item If the procedure in 3 isn't supported, use the PyTorch C++ API\footnote{https://pytorch.org/cppdocs/} to rewrite the model (only the torch.nn.Module\footnote{https://pytorch.org/docs/stable/nn.html\#module} subclass) and wrap it in a Caffe2 operator.
\item Publish the model to the production-grade Caffe2 prediction service and start serving live traffic
\end{enumerate}

\subsection{Benchmarks}

\begin{table}[h]
\begin{tabular}{|l|l|l|l|l|}
\hline
Model                       & Implementation              & P50            & P90            & P99            \\ \hline
\multirow{2}{*}{JointBLSTM} & PyTorch                     & 34.08          & 47.23          & 64.94          \\ 
                            & \textbf{Exported to Caffe2} & \textbf{19.65} & \textbf{24.69} & \textbf{30.21} \\ \hline
\multirow{2}{*}{RNNG}       & PyTorch                     & 19.74          & 28.53          & 36.37          \\ 
                            & \textbf{PyTorch C++}        & \textbf{18.73} & \textbf{25.47} & \textbf{32.63} \\ \hline
\end{tabular}
\caption{Latency Comparison (in milliseconds, smaller is better) of Python and C++ implementations of PyText models}
\label{table:benchmark}
\end{table}

We compared the performance of Python and C++ models (either directly exported to Caffe2 or re-written with the PyTorch C++ API\footnote{\label{ossnote}Currently not a part of PyText's open-source repository}) on an intent-slot detection task. We note that porting to C++ gave significant latency boosts (Table \ref{table:benchmark}) for the JointBLSTM model and a slight boost for the RNNG model. The latter is still valuable though, since the highly performant production serving infrastructure in many companies don't support Python code.

The experiments were performed on a CPU-only machine with 48 Intel Xeon E5-2680 processors clocked at 2.5GHz, with 251 GB RAM and CentOS 7.5. The C++ code was compiled with gcc -O3.

\subsection{Production Challenges}
\subsubsection{Data pre-processing}
One limitation of PyTorch is that it doesn't support string tensors; which means that any kind of string manipulation and indexing needs to happen outside the model. This is easy during training, but makes productionization of the model tricky. We addressed this by writing a featurization library in C++\cref{ossnote}. This is accessible during training via Pybind\footnote{https://github.com/pybind/pybind11} and at inference as part of the runtime services suite shown in Figure \ref{fig:architecture}. This library preprocesses the raw input by performing tasks like --
\begin{itemize}
\item Text tokenization and normalization
\item Mapping characters to IDs for character-based models
\item Perform token alignments for gazetteer features
\end{itemize}

By sharing the featurization code across training and inference we ensure data consistency in the different stages of the model.


\tikzstyle{doc}=[%
draw,
thick,
align=center,
color=black,
shape=document,
minimum width=1cm,
minimum height=1cm,
shape=document,
inner sep=2ex,
]

\begin{figure}[!h]
\begin{center}
\small
\begin{tikzpicture}[node distance = 1.25cm, auto]
    \node [block3, text width=3cm, minimum height=6cm, text height=-5 cm] (training) {Training};
    \node [draw,rectangle split, rectangle split horizontal,rectangle split parts=3,minimum height=1cm, above of=training, node distance=4.25cm, xshift=2.25cm, fill=BrickRed!15] (featurizer) {\nodepart{two}\shortstack{Featurizer Library (C++)}};
    \node [doc, below of=training, node distance=-1.5cm, fill=yellow!5] (rawdata) {Raw Data};
    \node [block4, below of=rawdata] (datahandler) {Data Handler};
    \node [block4, below of=datahandler] (trainer) {Trainer};
    \node [block4, below of=trainer] (exporter) {Exporter};
    \node [cloud1, below of=exporter, node distance=2cm, text width=1cm] (caffe2) {Caffe2 Model};
    \node [block3, text width=3cm, minimum height=6cm, right of=training, node distance=4.5cm, text height=-5 cm] (testing) {Inference};
    \node [doc, below of=testing, node distance=-1.5cm, fill=yellow!5] (rawdata2) {Raw Data};
    \node [block4, below of=rawdata2] (preprocess) {Data Preprocessor};
    \node [block4, below of=preprocess] (predictor) {Predictor};
    \node [doc, below of=predictor, fill=YellowGreen!15, node distance=3.25cm] (predictions) {Predictions};
    \path [line] ($(featurizer.south)+(-0.2,0)$) |- (datahandler);
    \path [line] ($(featurizer.south)+(0.2,0)$) |- (preprocess);
    \path [line] (rawdata) -- (datahandler);
    \path [line] (datahandler) -- (trainer);
    \path [line] (trainer) -- (exporter);
    \path [line] (exporter) -- (caffe2.north);
    \path [line] (caffe2.east) --++(1,0) --++(0,3.25) --++(0.85,0) (predictor);
    \path [line] (rawdata2) -- (preprocess);
    \path [line] (preprocess) -- (predictor);
    \path [line] (predictor) -- (predictions);
\end{tikzpicture}
\caption{Training and Inference Workflow Architecture}
\label{fig:architecture}
\end{center}
\end{figure}
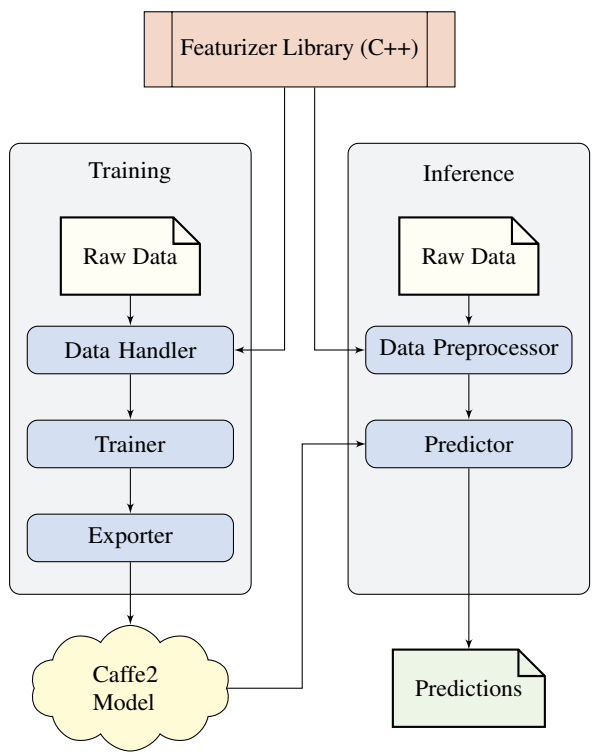

\subsubsection{Vocabulary management}
Another consequence of string tensors not being supported yet is that we can't maintain vocabularies inside the model. We explored two solutions to this --
\begin{itemize}
\item Maintain the vocabularies in the remote featurization service.
\item After exporting the model, post-process the resultant Caffe2 graph and prepend the vocabularies to the net
\end{itemize}
We ultimately opted for the second option since its non-trivial to maintain synchronization and versioning between training-time and test-time vocabularies, across different use cases and languages.

\section{Future Work}
Upcoming enhancements to PyText span multiple domains:
\begin{itemize}
\item \textbf{Modeling Capabilities:}
Adding support for advanced NLP models for more use cases, e.g.
\begin{itemize}
    \item Question answering, reading comprehension and summarization tasks
    \item Multilingual and language-agnostic tasks
\end{itemize}
\item \textbf{Performance Benchmarks and Improvements :} A core goal of PyText is to enable building highly scalable models, with can run with low latency and high throughput. We plan to invest in --
\begin{itemize}
\item Training speed -- by augmenting the current distributed-training support with lower precision computations support like fp16\footnote{https://en.wikipedia.org/wiki/Half-precision\_floating-point\_format}
\item Inference speed -- by benchmarking performance and tuning the model deployment for expected load patterns.
\end{itemize}
\item \textbf{Model Interpretability:}
We plan to add more tooling support for monitoring metrics and debugging model internals --
\begin{itemize}
\item Tensorboard\footnote{https://github.com/tensorflow/tensorboard} and Visdom \footnote{https://github.com/facebookresearch/visdom} integration for visualizing the different layers of the models and track evaluation metrics during training
\item Explore and implement different model explanation approaches, e.g LIME \footnote{https://github.com/marcotcr/lime} and SHAP \cite{DBLP:journals/corr/LundbergL17}
\end{itemize}
\item \textbf{Model Robustness:} Adversarial input, noise, and differences in grammar and syntax can often hurt model accuracy. To analyze and improve robustness against these perturbations, we plan to invest in adversarial training and data augmentation techniques.

\item \textbf{Mobile Deployment Support:} We utilize the optimized Caffe2 runtime engine to serve our models, and plan to leverage its optimization for mobile devices \footnote{https://caffe2.ai/docs/mobile-integration.html}, as well as support training light-weight models.

\end{itemize}
\section{Conclusion}
In this paper we presented PyText -- a new NLP modeling platform built on PyTorch. It blurs the boundaries between experiments and large scale deployment and makes it easy for both researchers and engineers to rapidly try out new modeling ideas and then productionize them. It does so by providing an extensible framework for adding new models and by defining a clear production workflow for rigorously evaluating and serving them. Using this framework and the processes defined here, we significantly reduced the time required for us to take models from research ideas to industrial-scale production.
\bibliography{pytext_paper}
\bibliographystyle{sysml2019}

\end{document}